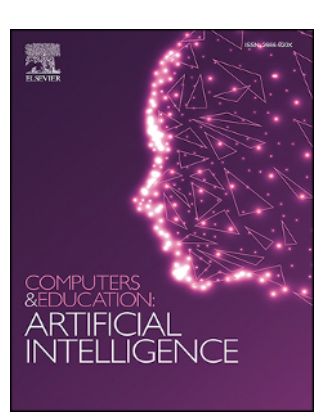

# The promise and limits of LLMs in constructing proofs and hints for logic problems in intelligent tutoring systems

Sutapa Dey Tithi*, Arun Kumar Ramesh, Clara DiMarco, Xiaoyi Tian, Nazia Alam, Kimia Fazeli, Tiffany Barnes

*Department of Computer Science, North Carolina State University, Raleigh, NC, USA*

ABSTRACT

Intelligent tutoring systems have demonstrated effectiveness in teaching formal propositional logic proofs, but their reliance on template-based explanations limits their ability to provide personalized student feedback. While large language models (LLMs) offer promising capabilities for dynamic feedback generation, they risk producing hallucinations or pedagogically unsound explanations. We evaluated the stepwise accuracy of LLMs in constructing multi-step symbolic logic proofs, comparing six prompting techniques across four state-of-the-art LLMs on 358 propositional logic problems. Results show that DeepSeek-V3 achieved superior performance with upto 86.7 % accuracy on stepwise proof construction and excelled particularly in simpler rules. We further used the best-performing LLM to generate explanatory hints for 1050 unique student problem-solving states from a logic ITS and evaluated them on 4 criteria with both an LLM grader and human expert ratings on a 20 % sample. Our analysis finds that LLM-generated hints were 75 % accurate and rated highly by human evaluators on consistency and clarity, but did not perform as well in explaining why the hint was provided or its larger context. Our results demonstrate that LLMs may be used to augment tutoring systems with logic tutoring hints, but those hints require additional modifications to ensure accuracy and pedagogical appropriateness.

## 1. Introduction

Intelligent tutoring systems (ITSs) serve as scalable alternatives to human tutoring (Kardan & Conati, 2015; Ueno & Miyazawa, 2017). These systems automatically evaluate student solutions and provide adaptive feedback, including next-step hints and explanations. Such personalized assistance has been shown to enhance student learning outcomes (VanLehn, 2011). Although various data-driven methods have been shown to save time and resources by reducing the need for an expert to construct hints, they rely on the quantity and quality of training data. Moreover, the hints in these systems are often template-based with limited adaptation (Stamper et al., 2024). In this context, large language models (LLMs) can offer opportunities to augment ITS capabilities through scalable feedback generation (Tian et al., 2024). However, integrating LLMs into ITSs faces challenges: LLMs may hallucinate incorrect information (Wang et al., 2024), which risks misleading students and reinforcing misconceptions (Nye et al., 2023). LLMs may also generate correct answers without adhering to sound instructional principles, potentially limiting their effectiveness in fostering deep learning (Aleven & Koedinger, 2002; Koedinger et al., 2012). Complicating matters for integrating LLM-based help in tutoring systems, novices may lack metacognitive skills to identify what and when to ask for help (help avoidance), and they could abuse the system by asking for direct answers or overly frequent hints (help abuse) (Roll et al., 2011). Thus, the quality of solutions, adherence to pedagogical principles, and adaptation based on metacognitive needs are as crucial as accuracy for help in learning environments (Holstein et al., 2019).

Formal logic proofs are essential in STEM education. Several studies have evaluated LLM performance in solving logic problems and handling natural language logical reasoning tasks (Parmar et al., 2024; Saparov & He, 2022). However, LLMs' ability to handle complex, multi-step symbolic logic proofs within pedagogical contexts is largely unexplored. In this work, we evaluate how well LLMs construct multi-step symbolic propositional logic proofs and next-step natural language hints in an intelligent logic tutor. We compared six prompting techniques across four state-of-the-art LLMs on 358 propositional logic problems, leveraging advanced prompt engineering to incorporate learning science theories. The quality of hints and explanations from the best-performing LLM was

* Corresponding author.
*Email address:* stithi@ncsu.edu (S.D. Tithi).

assessed by a rubric-based LLM grader and 20 % were also rated by human experts on a 4-dimensional rubric. We investigated the following research questions:

**RQ1** To what extent can LLMs construct accurate step-by-step solutions for propositional logic (PL) proofs within a logic tutor?

- RQ1a: How do prompting strategies influence proof construction performance?
- RQ1b: How does performance vary across a selection of four open-source and proprietary LLM models?

**RQ2** How well does the most accurate LLM generate pedagogically viable hints and explanations?

- RQ2a: How does hint correctness vary across logic rules of differing complexity?
- RQ2b: How do domain experts assess the pedagogical quality of LLM-generated explanations in terms of consistency, clarity, justification, and subgoaling?

## 2. Related work

**Intelligent Tutoring Systems and Automated Feedback.** ITSs have established their effectiveness in promoting student learning through personalization across multiple fields, including mathematics (Anderson et al., 1995; Canfield, 2001; Koedinger et al., 1997; Ritter et al., 2007), probability (Chi & VanLehn, 2010; Sedlmeier, 1997), logic (Maniktala & Barnes, 2020), and programming languages (Anderson et al., 1995; Holland et al., 2009; Price, Dong & Lipovac, 2017). The evolution of these systems has led to various data-driven techniques for providing adaptive support. Murray (2003) demonstrated that data-driven assistance can efficiently generate hints while reducing expert involvement. This approach was further advanced by Barnes et al. through the Hint Factory, a method for generating next-step hints in propositional logic (Barnes et al., 2008). Further research has confirmed that both on-demand and unsolicited hints can significantly enhance student learning when appropriately implemented (Barnes et al., 2008; Maniktala & Barnes, 2020; Stamper et al., 2013). Despite the popularity and benefits of data-driven hints in ITSs, the current approaches have some limitations. Many systems provide immediate next-step hints without offering conceptual explanations (Price, Dong & Lipovac, 2017; Rivers & Koedinger, 2017). Marwan et al. showed that novices perceived hints with explanations as significantly more relevant and interpretable than those without explanations in a block-based programming environment, and hints without conceptual explanations can decrease students' trust in their helpfulness (Marwan et al., 2019). While incorporating textual explanations alongside procedural hints has been shown to be useful, generating these explanations traditionally requires extensive manual effort.

**Large Language Models in Logic.** Recent applications demonstrate LLMs' effectiveness in providing conversational support and personalized instruction across various educational contexts (Roest et al., 2024; Taneja et al., 2024; Venugopalan et al., 2024). Aashish et al. showed that students in an introductory programming course utilized a custom GPT-4 AI tool for assignment help, and students primarily asked AI assistance for conceptual understanding rather than complete solutions (Ghimire & Edwards, 2024). Systems like HypoCompass have demonstrated that LLM-augmented ITS can generate high-quality training materials and significantly improve student performance (Ma et al., 2024). However, the application of LLMs to formal logic instruction presents unique challenges. While foundational datasets like ProofWriter (Tafjord et al., 2020) enable multi-hop proofs and FOLIO (Han et al., 2022) introduces complex first-order logic (FOL) expressions, existing benchmarks remain narrowly scoped. ProntoQA (Saparov & He, 2022) focuses on just one FOL rule (modus ponens). LogicBench (Parmar et al., 2024) covers a larger set of inference rules, and it reveals critical performance gaps even in state-of-the-art models like GPT-4. Moreover, performance can decrease for complex operations, with accuracy dropping below 45 % for proofs requiring three or more steps (Saparov et al., 2023). Prompt engineering techniques like Chain-of-Thought (CoT) have improved LLMs' reasoning capabilities, but ensuring the faithfulness of generated reasoning chains remains challenging (Liu, Xu et al., 2024). Careful prompt engineering is also important in ensuring the effective utilization of LLMs in educational contexts (Stamper et al., 2024). These approaches predominantly address LLM performance on natural language logic problems or isolated inference rules, with limited work exploring multi-step symbolic propositional logic proofs. Additionally, these LLMs are evaluated using traditional metrics such as accuracy and coverage (Phung et al., 2023, 2024); they are rarely assessed for the suitability of their explanations for educational contexts. Our work bridges these gaps by investigating LLMs' capabilities in multi-step symbolic logic proof construction and explanatory hint generation, with the goal of combining the pedagogical rigor of traditional ITS with the natural language capabilities of LLMs. We extend traditional evaluation metrics to include consistency (the what), clarity (the readability), justification (the why), and subgoaling (describing the why within a larger context), as described below, providing a more comprehensive assessment of LLM-generated hint quality in logic education.

## 3. Methodology

**Proof Construction.** We selected two datasets to evaluate LLMs' performance in proof construction: (1) The Symbolic Propositional Logic (*SPL*) dataset contains 310 multi-step problems. These problems were originally in natural language form, then Gottlieb et al. systematically translated them into symbolic PL representations (Gottlieb, 2024). (2) The Logic Tutor dataset (*LT*) contains 48 PL problems of varying difficulty. These problems are designed specifically for educational purposes, with each proof typically requiring 12–15 steps to complete. In both SPL and LT datasets, each problem contains the given premises and a conclusion, which are fed into LLMs using carefully engineered prompts. LLMs output the complete step-by-step proof along with a structured solution in JSON format.

**Logic Tutor Problem Solving State (PSS) Dataset Extraction.** To enable realistic testing of next-step hint generation, a dataset was extracted from interaction logs of all solution attempts to 18 problems each by 30 students using the logic tutor for homework in Fall 2024. Students were randomly selected from all CS majors at a US public university, whose 2021–22 graduating class demographics comprised 83 % men and 17 % women; with 58 % white, 18.5 % Asian, 3 % Hispanic/Latin, 2 % Black/African American, 9 % other races with the remaining 9.5 % having international student status for whom race/ethnicity information was not available. IRB approval was obtained, and only authorized researchers could access participant data. We developed an automated pipeline to convert the interaction logs into a textual representation of student problem solving states (PSS) (snapshots of problem solving after each step) (Barnes & Stamper, 2008) (Fig. 1). These PSS representations were fed into our final LLM framework using carefully engineered prompts to generate targeted next-step hints along with conceptual explanations for each PSS (Fig. 1).

**Intelligent Logic Tutor.** Our logic tutor presents propositional logic (PL) problems to students in a graphical representation. Each problem shows a set of given statements at the top of the workspace and a conclusion to be derived at the bottom (Fig. 2). Students solve the problems by iteratively deriving new logic statements until they arrive at the conclusion. The tutor divides the problems into three sections: pretest, training, and posttest. Students' prior knowledge is measured in the pretest section. The training section consists of five ordered levels with increasing difficulty. This is the only section where students can request and receive hints. Finally, students complete a posttest section consisting of six problems without any tutor assistance. The tutor uses a data-driven method to generate next-step hints with template-based explanations to guide students given their current progress.

**Student Progress**

**Final Conclusion:** $J \vee K$
**Derivation Steps:**
1. $F \rightarrow (G \wedge \neg H)$, Given
2. $I \wedge F$, Given
3. $H \vee J$, Given
4. $\neg F \vee (G \wedge \neg H)$, Derived from (1) using *Impl*

**LLM-Generated Hint**

**Hint:** Derive $F$ using *Simp* on line 2.
**Explanation:**
Line 2 ($I \wedge F$) contains $F$, which is important for simplifying other statements. By deriving $F$ from $I \wedge F$, we can use it to simplify line 4 ($\neg F \vee (G \wedge \neg H)$) and move closer to the conclusion.

**Tutor-Generated Hint**

**Hint:** Click $I \wedge F$ and click on rule Simplification to derive $F$.

**Fig. 1.** Example of a student problem-solving state and generated hints. The top box presents the student's derivation steps, while the bottom two boxes present the LLM- and tutor-generated hints.

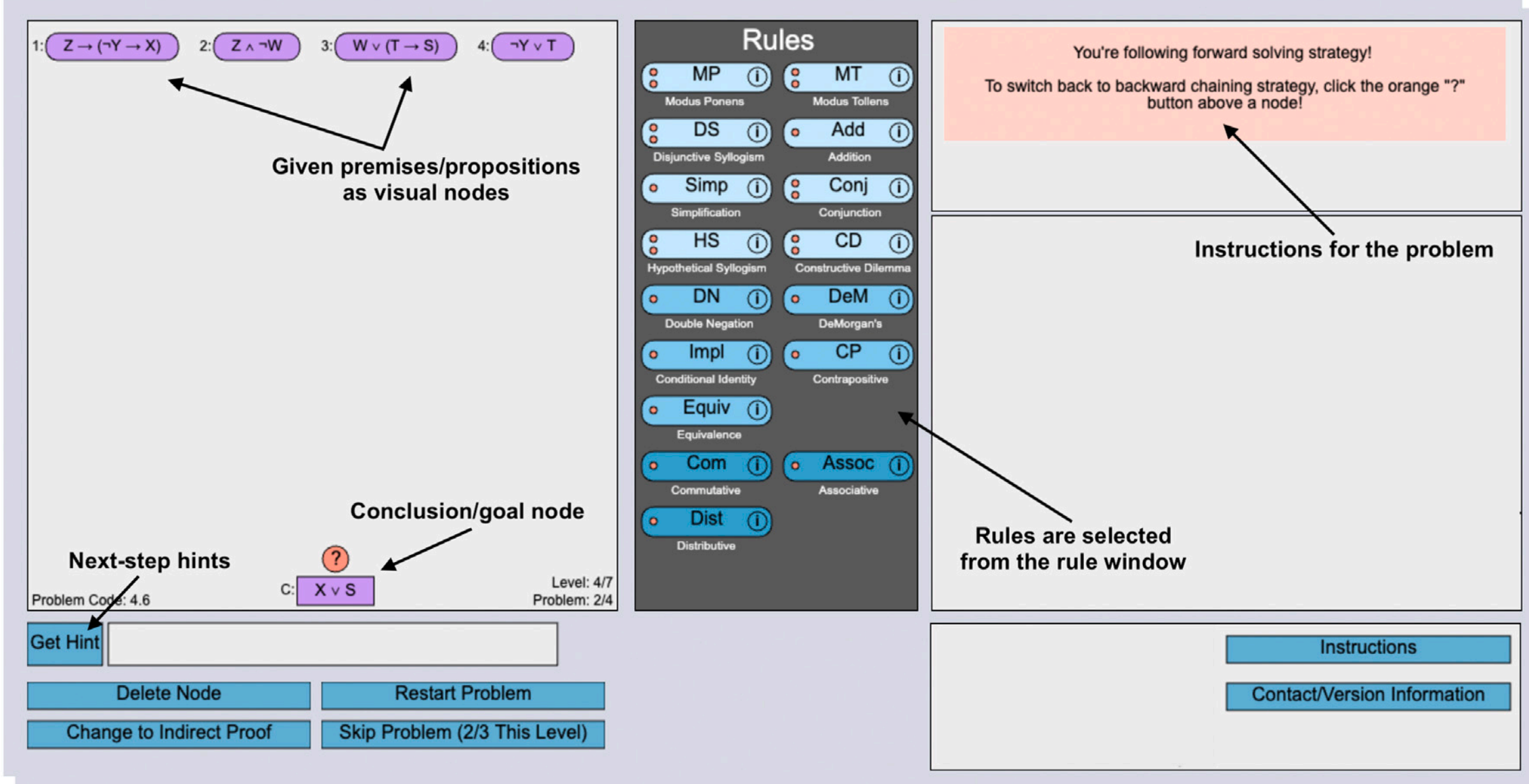

**Fig. 2.** The full interface of the logic tutor, showing the given statements at the top of the workspace and a conclusion to be derived at the bottom. The middle pane contains the rule buttons.

### *3.1. Model selection and prompt engineering*

We evaluated four state-of-the-art LLMs: Llama-3-70B-Instruct (AI@Meta, 2024), Gemini-Pro (Anil et al., 2023), GPT-4o (Achiam et al., 2023), and DeepSeek-V3 (Liu, Feng et al., 2024). We set the temperature parameter to 0.1 for all LLMs to limit the randomness and variability of responses and improve accuracy. We opted not to use the LLMs, such as o1 or DeepSeek-Reasoner, as they currently require significant computational resources and are not yet practical for deployment in a classroom setting. Each prompt is structured consistently (Fig. 3) and is designed to incorporate evidence-based learning science principles (Stamper et al., 2024). Our prompt engineering framework implements six increasingly advanced techniques, building upon Zero-Shot (ZS) and Few-Shot (FS) prompting (Brown et al., 2020) with Chain-of-Thought (CoT) reasoning (Wei et al., 2022). The techniques are summarized in Table 1.

1. **Zero Shot (ZS)** Zero-shot prompting elicits step-by-step solutions and output expectations without providing any training examples. Thus, it relies on the model's inherent reasoning and problem-solving capabilities, testing its capacity to generalize with no prior guidance.
2. **Few Shot Plan and Solve (FS_PlanAndSolve)** This technique instructs the LLM to decompose a problem into subgoals and address each subgoal one by one. Two training examples are included in the prompt to demonstrate this approach, with the goal of helping the model learn to break its problem-solving process into smaller intermediate subgoals. This problem decomposition–recomposition strategy promotes logical consistency across complex tasks.
3. **Few Shot Chain of Thought (FS_CoT)** Chain-of-thought (CoT) prompting provides the model with examples that demonstrate a series of intermediate reasoning steps. Through these exemplars, the model may learn to articulate its own reasoning more transparently. Previous studies have shown that CoT significantly improves the reasoning abilities of LLMs (Wei et al., 2022).

**Abstract Prompt Template for Proof/Hint Generation**

**Role Assignement & Context**
*(e.g., "You are an intelligent tutor in propositional logic...")*

**Instructions**
1. Step-by-step proofs, where each step should have derived node, parents, rule applied
3. Acceptable rules: <allowed rules in the tutor>
4. Final "structured solution" in JSON-like format

**Output Expectation**
- A detailed step-by-step derivation followed by a JSON-like "structured solution."
- A detailed step-by-step derivation, a next-step hint followed by a conceptual explanation

**Few-Shot Examples**
- Demonstrates prompt specific proof strategies, step-by-step derivations and the final JSON-like output.
- Demonstrates a step-by-step proof, one-line next-step hint, a conceptual explanation of the hint

**User Prompt**
References givens, conclusion and/or current progress and instructions to produce a formal proof/hint.

**Fig. 3.** Abstract prompt template incorporating context, instructions, output expectations, examples, and user prompts.

**Table 1**
Summarized descriptions of six prompting techniques.

| Technique | Title & description |
| --- | --- |
| 1. ZS | **Zero Shot:** Demonstrates step-by-step solutions without training examples, relying on the model's inherent reasoning capabilities. |
| 2. FS_PlanAndSolve | **Few Shot Plan and Solve:** Instructs the LLM to decompose problems into subgoals, demonstrated through two training examples aiming for better logical consistency. |
| 3. FS_CoT | **Few Shot Chain of Thought:** Provides examples demonstrating intermediate reasoning steps, enabling the model to articulate transparent reasoning. |
| 4. FS_L_DCoT | **Few Shot Logical Dual CoT:** Integrates both direct and indirect proof strategies, demonstrating alternate logical paths and cross-verification. |
| 5. FS_BL_DCoT | **Few Shot Bidirectional Logical Dual CoT:** Combines forward chaining (premises to conclusion) with backward chaining (conclusion to premises) for additional validation. |
| 6. FS_ToT_CoT | **Few Shot Tree of Thoughts CoT:** Uses multiple reasoning "threads" where experts collaboratively debate, forming a tree-like structure that prunes failed reasoning paths and validates successful ones. |

4. **Few Shot Logical Dual Chain of Thought (FS_L_DCoT)** Building on the CoT concept, this approach integrates both direct and indirect proof strategies. This technique provides the LLM with examples that demonstrate how to consider alternative logical paths and cross-verify them. By encouraging the exploration of multiple reasoning paths, this approach reduces the likelihood of overlooking potential mistakes or alternative solutions.
5. **Few Shot Bidirectional Logical Dual Chain of Thought (FS_BL _DCoT)** This strategy augments the Dual CoT by combining forward chaining (from premises to conclusion) with backward chaining (from conclusion back to premises). By working in both directions, the model may gain an additional layer of validation and error detection.
6. **Few Shot Tree of Thoughts Chain of Thought (FS_ToT_CoT)** In this approach, the solution path branches out across multiple reasoning "threads" or experts, each contributing a step-by-step proof. We added examples in the prompt where three experts collaboratively debated each step, acknowledged and corrected mistakes, and refined their step derivations. This collaborative interplay forms a tree-like structure of possible solution paths: if a line of reasoning fails, it is pruned when the expert acknowledges their mistake; successful lines continue to expand. This approach is intended to ensure that alternative proofs are considered and validated.

### 3.2. Data splits

We used the first two examples that students see in the tutor (as shown in Appendix A) for few-shot prompting of every LLM to provide appropriate logic proof structures. We set the LLM temperature parameter to 0.1 to achieve consistency but allow for some randomness. For each dataset, we split the data into validation (35 %) and test (65 %) sets using a fixed random seed (42) to ensure reproducibility. To refine the LLM prompts, we iteratively refined the prompt templates based on the *validation* set performance guided by two primary criteria: (1) maximizing accuracy, and (2) minimizing syntax errors detected by our automated logic checker (i.e., solutions that could not be parsed). The iteration process was stopped when accuracy plateaued and no further improvements were observed across consecutive refinement attempts. To mitigate the risk of overfitting to the validation set, we allocated the majority (65 %) of the data to the test set. Once the prompt template was finalized, we ran the models only on the *test* set and reported the final results.

### 3.3. Evaluation framework

The built-in logic checker in the logic tutor automatically verified the correctness of all LLM-generated proof steps and hints, ensuring relevance within the tutor context. We conducted a qualitative evaluation of LLM-generated hints and explanations by two experts with 2–4 years of experience as teaching assistants and researchers in logic and ITS.

The evaluation rubrics (Table 2) were comprised of four key metrics: consistency, clarity, justification, and subgoaling. To ensure the suitability of the metrics for our tutor context, our two domain experts collaborated and iteratively refined them until a consensus was reached with a course instructor expert. These metrics capture essential qualities of effective hint explanations proposed in previous work (Daheim et al., 2024; Jeuring et al., 2022; Macina et al., 2023; Maurya et al., 2024; Roest et al., 2024; Tack & Piech, 2022). First, high-quality hints must maintain logical *consistency* with both the student's current proof state and the domain rules permitted by the tutor (Macina et al., 2023). Second, hints should provide *clarity* through clear, actionable guidance without vagueness or extraneous details, enabling students to understand their next steps (Daheim et al., 2024; Kotalwar et al., 2024). Third, effective hints incorporate *justification* or rationale to deepen student understanding of the underlying reasoning (Jeuring et al., 2022, Tack & Piech, 2022). Finally, hint explanations benefit from including *subgoals* as high-level guidance to help students develop an understanding of the more general problem-solving skills beyond immediate procedural steps (Cucuiat & Waite, 2024; Morrison et al., 2016; Shabrina et al., 2023; Tithi et al., 2025).

**Table 2**
Evaluation Criteria for LLM-generated hint explanations.

| Criteria | Definition |
| --- | --- |
| Consistency | Does the hint follow the tutor's domain rules and align with the student's current proof state? |
| Clarity | Is the hint clearly written and free of irrelevant or excessive details? |
| Justification | Does the hint provide a rationale or reasoning for why the next step was suggested? |
| Subgoaling | Does the hint provide a larger context (without giving away the full solution) explaining how the step will help work on other statements? |

We used stratified random sampling, with stratification done by problem to ensure problem diversity. For each problem, we sampled 20 % from the set of unique solution states, using a fixed random seed (42) for reproducibility. The hints and explanations for the sampled solution states were rated on each rubric on a scale from 1 to 4 (with 4 being the best). We used Spearman correlation ($\rho$) and Quadratic Weighted Cohen's Kappa ($QWK$) to obtain inter-rater agreement on the assessment dimensions, which are common reliability measurements between two raters (Naismith et al., 2023; Tian et al., 2024). We also used a rubric-based LLM grader to understand the potential of using it as an automated evaluator.

## 4. Results

### 4.1. RQ1: evaluating logic proof construction performance

We evaluated the performance of four LLMs (Llama-3, Gemini-Pro, GPT-4o, and DeepSeek-V3) using six prompting techniques on the two logic tutor (LT) and symbolic propositional logic (SPL) datasets. DeepSeek-V3 demonstrated superior accuracy in most prompt configurations, with the best FS_ToT_CoT prompts achieving 85.0 % and 86.7 % accuracy for the steps generated for full solutions of LT and SPL problems, respectively (Table 3). GPT-4o also exhibited strong performance, particularly on SPL with FS_CoT prompts (83.6 %). Gemini-Pro showed mixed results, while Llama-3 generally underperformed. All zero-shot prompted models underperformed, indicating the importance of in-context prompting in the integration of LLMs into ITSs.

To differentiate the capabilities of the LLMs, we explored the complexity of each problem, as measured by the lengths of the statements combined to derive new steps. We hypothesized that longer parent statements would be more likely to lead to reasoning errors (incorrect steps), as they typically do for students. In Table 4, we show the mean lengths of the parent statements for correct steps (in column 'C') and for incorrect steps (in column 'Inc') for four LLMs (Llama-3, Gemini-Pro, GPT-4o, and DeepSeek-V3) in six prompting techniques on both LT and SPL datasets combined.[1] We compared the lengths of parent statements between correct and incorrect steps, and for each LLM, the lengths of the parent statements were significantly longer for incorrect versus correct steps with $p < 0.001$ in each comparison. This suggests longer parent statements may introduce ambiguity, leading to more incorrect steps.

**Rule-Specific Performance Analysis.** For all the FS-based prompting techniques and the GPT-4o and DeepSeek-V3 models, we calculated the average accuracy across different logical rules. As shown in Table 5, DeepSeek-V3 and GPT-4o both achieved high accuracy on basic rules (Com, Add, Conj, Simp) while DeepSeek-V3 maintained stronger performance on the rules MP, DeM, MT, HS, DS and Contra. The performance of GPT-4o declined with higher rule complexity (as measured by number of operators including negations). The most challenging rules were Contra, CP, DN, and Impl, although DeepSeek-V3 still outperformed GPT-4o.

### 4.2. RQ2a: evaluating next-step hint generation performance

Based on RQ1, DeepSeek-V3 performed better than all other LLMs. While more complex prompting like bidirectional dual CoT and tree-of-thoughts CoT improved accuracy, their lengthier and more complex prompts appeared difficult to adapt for the hint generation task. FS_CoT achieved comparable accuracy with simpler prompting, making it our preferred choice. Using DeepSeek-V3 with FS_CoT, we generated next-step hints and conceptual explanations for the LT PSS dataset containing 1050 unique student problem-solving states. Each state represents a snapshot of a student's progress, containing all steps completed at that point, and transformed from a graphical into a natural language-based form. DeepSeek-V3 with FS_CoT was used to generate the next best step hint for each state. A modified version of the LT logic checker was used to evaluate correctness, with any hint considered to be incorrect if the step was logically incorrect, the suggested step was already present in the student solution, or if the parent steps were not yet derived. Overall, 75 % of the resulting hints were rated as correct/accurate. Accuracy across the specific logical rules is shown in Fig. 4. The model struggled most with complex rules like DS, Impl, Add, DN, MT, DeM, CP. With the exception of Addition, all of these rules handle negations, introducing complexity into the application of rules. Addition introduces a new variable into a proof, which adds another type of complexity.

We compared the performance of DeepSeek-V3 to our logic tutor's data-driven hint mechanism, which utilizes historical data to identify optimal next steps. DeepSeek-V3 generated a higher number of unique hints per problem on average (M = 4.5) compared to the tutor (M = 3.0). The overall hint accuracy of 75 % is promising. However, accuracy varied across the tutor levels. While the LLM achieved 87.9 % accuracy in the first training level, the number of unique hints increased, and accuracy declined across the 5 training levels. Since difficulty and the number of logic rules needed increase at each level, the size of the potential solution space expands. This larger solution space may have led to lower hint accuracy and the generation of more unique hints by DeepSeek-V3.

**Table 3**
Performance of different LLMs in different zero-shot and few-shot settings, measured by accuracy on two datasets: LT and SPL. **Boldface** indicates the best accuracy score for each prompt technique.

| Prompt Technique | Llama-3 | | Gemini-Pro | | GPT-4o | | DeepSeek-V3 | |
|---|---|---|---|---|---|---|---|---|
| | LT | SPL | LT | SPL | LT | SPL | LT | SPL |
| 1. ZS | 24.4 | 34.1 | 37.7 | 40.5 | 67.9 | 68.3 | **75.7** | 72.2 |
| 2. FS_CoT | 65.7 | 62.6 | 59.0 | 67.0 | 76.9 | 83.6 | 83.7 | **84.4** |
| 3. FS_PlanAndSolve | 56.7 | 60.2 | 49.8 | 50.3 | 75.5 | 79.9 | 79.0 | **83.0** |
| 4. FS_BL_DCoT | 49.1 | 52.3 | 58.5 | 67.4 | 76.7 | 79.8 | 84.4 | **84.9** |
| 5. FS_L_DCoT | 63.8 | 64.6 | 54.9 | 50.3 | 73.5 | 75.5 | 84.4 | **84.4** |
| 6. FS_ToT_CoT | 59.1 | 56.1 | 51.7 | 60.4 | 72.7 | 82.4 | 85.0 | **86.7** |

**Table 4**
Mean *Parent length* for Correct vs Incorrect steps across models and prompt techniques. Within each model, the shaded 'C' column reports the mean for correct steps; the 'Inc' column reports the mean for incorrect steps.

| Prompt Technique | Llama-3 | | Gemini-Pro | | GPT-4o | | DeepSeek-V3 | |
|---|---|---|---|---|---|---|---|---|
| | C | Inc | C | Inc | C | Inc | C | Inc |
| 1. ZS | 2.3 | 6.0 | 2.8 | 7.0 | 3.9 | 6.5 | 4.0 | 8.7 |
| 2. FS_CoT | 4.4 | 6.9 | 3.9 | 6.5 | 4.9 | 8.0 | 4.8 | 8.1 |
| 3. FS_PlanAndSolve | 4.3 | 8.1 | 3.6 | 8.2 | 4.6 | 7.5 | 4.8 | 9.8 |
| 4. FS_BL_DCoT | 4.1 | 8.7 | 4.1 | 7.1 | 4.8 | 6.3 | 4.7 | 7.4 |
| 5. FS_L_DCoT | 4.4 | 7.1 | 4.0 | 6.9 | 4.6 | 6.9 | 4.8 | 9.0 |
| 6. FS_ToT_CoT | 4.3 | 6.5 | 3.6 | 6.0 | 4.6 | 6.9 | 4.8 | 7.5 |

**Table 5**
Rule accuracy for comparison of DeepSeek-V3 and GPT-4o across all prompting techniques.

| Rule | Detail | DeepSeek-V3 | GPT-4o |
|---|---|---|---|
| **Com**mutation | $(p \lor q) \implies (q \lor p)$ | 99.23 | 89.28 |
| **Add**ition | $p \implies (p \lor q)$ | 95.46 | 87.45 |
| **Simp**lification | $(p \land q \implies p)$ | 92.16 | 83.34 |
| **Conj**unction | $(p, q) \implies (p \land q)$ | 90.76 | 90.00 |
| Modus Ponens (**MP**) | $((p \to q) \land p) \implies q$ | 89.17 | 82.26 |
| DeMorgan's (**DeM**) | $\neg(p \land q) \implies (\neg p \lor \neg q)$ | 87.01 | 76.97 |
| Modus Tollens (**MT**) | $((p \to q) \land \neg q) \implies \neg p$ | 86.80 | 74.68 |
| Hyp. Syll. (**HS**) | $((p \to q) \land (q \to r) \implies (p \to r)$ | 86.35 | 75.83 |
| Disj. Syll. (**DS**) | $((p \lor q) \land \neg p) \implies q$ | 78.06 | 62.31 |
| **Contra**diction | $(p \land \neg p) \iff 0$ | 76.67 | 38.24 |
| **Impl**ication | $(p \to q) \iff (\neg p \lor q)$ | 72.26 | 68.52 |
| Double Negation (**DN**) | $\neg\neg p \implies p$ | 69.20 | 67.05 |
| Contrapositive (**CP**) | $(p \to q) \implies (\neg q \to \neg p)$ | 64.07 | 45.77 |

[1] We present the mean lengths of the parent statements for correct steps and incorrect steps for each technique after combining the two datasets, LT and SPL. The results for each dataset separately yielded similar results.

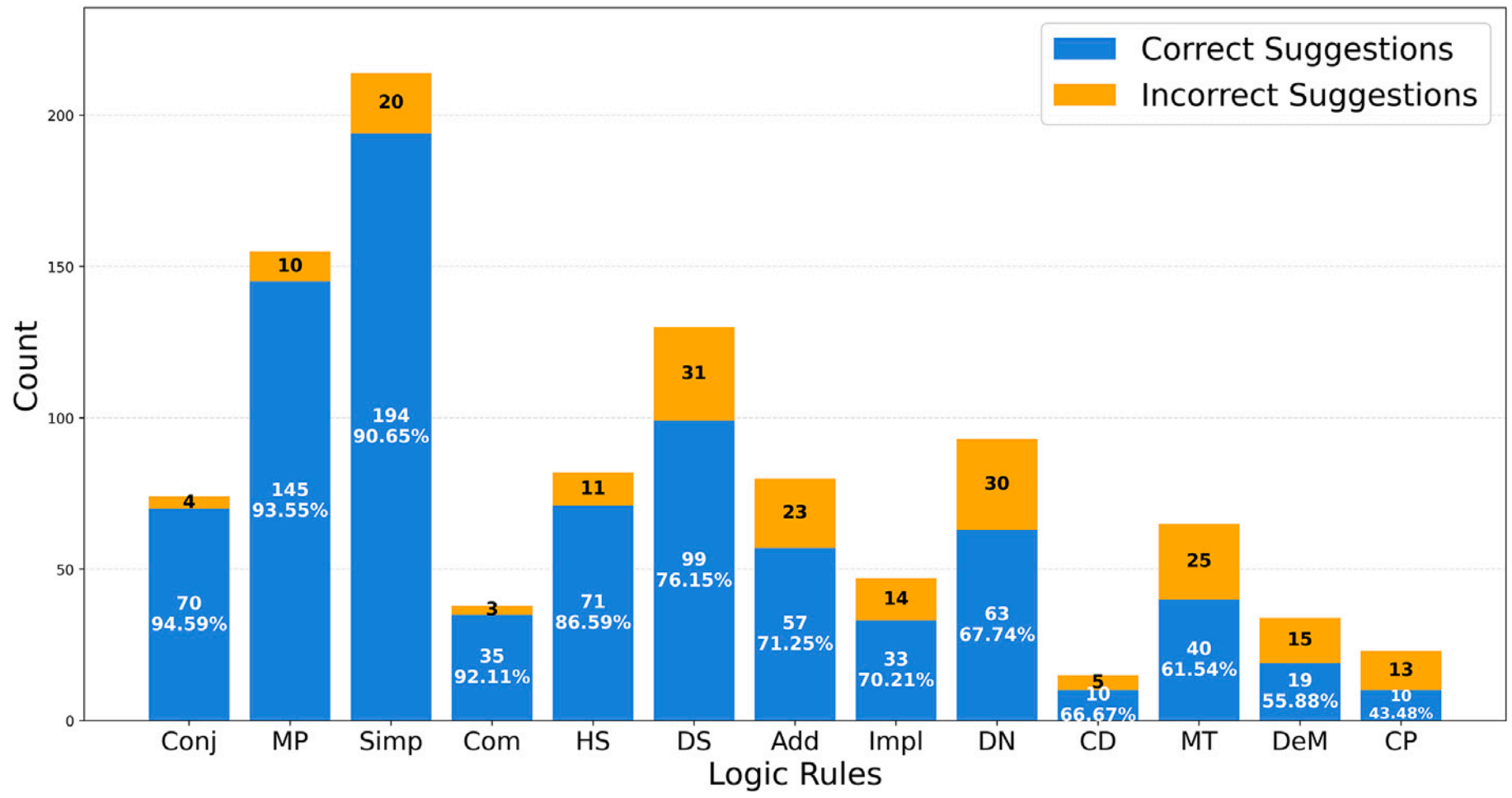

**Fig. 4.** The distribution of correct and incorrect LLM-generated hints for each rule.

**Table 6**
Inter-rater agreement metrics for evaluation of 210 LLM-generated explanations. ($\rho$ = Spearman's Correlation, $QWK$ = Quadratic Cohen's Kappa. $^{\dagger}p >$ Bonferroni$_{\text{threshold}}$).

| Dimension | Human-Human | | Avg rating (Human) | Human-LLM | | Avg Rating (LLM) |
|---|---|---|---|---|---|---|
| | $\rho$ | $QWK$ | | $\rho$ | $QWK$ | |
| Consistency | 0.90 | 0.85 | 3.9 | 0.59 | 0.30 | 3.81 |
| Clarity | 0.93 | 0.92 | 3.09 | 0.54 | 0.56 | 2.92 |
| Justification | 0.92 | 0.92 | 2.24 | 0.55 | 0.59 | 2.39 |
| Subgoaling | 0.77 | 0.79 | 1.66 | **0.08**$^{\dagger}$ | **0.16**$^{\dagger}$ | 2.47 |

### 4.3. RQ2b: evaluating next-step hint explanations

Two expert graders independently applied the evaluation rubrics (Table 2) to a sample of 210 LLM-generated conceptual explanations. Table 6 shows a strong inter-rater agreement for the expert human graders across all dimensions. The Spearman correlation ($\rho$) ranged from 0.77 to 0.93, and the $QWK$ values ranged from 0.79 to 0.92, suggesting near-perfect reliability (McHugh, 2012). These high agreement metrics suggest that our evaluation criteria were well-defined and consistently applied by the expert human raters. The sampled LLM explanations were highly rated for consistency and clarity, with human-rater scores of 3.90 and 3.09 out of 4, respectively. However, the system showed room for improvement in other areas, with expert human rater scores of 2.24 for justification and 1.66 for subgoaling. We applied an LLM grader using the same rubrics to the sample of explanations and compared their rubric scores with the expert human ratings. While the LLM grader assigned consistency and clarity scores comparable to human scores, it overestimated human scores in justification and subgoaling. However, the rubric-based LLM grader exhibited weaker agreement with human graders, particularly in subgoaling, where the average LLM subgoaling score was 2.47 as opposed to the human subgoaling score of 1.66.

## 5. Discussion

Logic rules can be considered basic or complex based on their use of negation. Basic rules such as Simplification or Commutative involve direct and single-step operations (e.g., $(p \vee q) \implies (q \vee p)$). On the other hand, the rules involving negation (e.g., DeMorgans) cause challenges for people and for LLMs. The implication operator $\rightarrow$ inherently includes negation as part of its definition ($(p \rightarrow q) \iff (\neg p \vee q)$). Statements that combine both negation and implication increase the complexity even more, such as Modus Tollens and Contrapositive ($(p \rightarrow q) \implies (\neg q \rightarrow \neg p)$). Johnson et al. argued that rules involving negation require working memory to track inverted truth values across multiple steps, thus increasing cognitive load (Johnson-Laird, 2001). Anderson et al. reported that humans exhibit slower response times and higher error rates when processing negated statements like $\neg(A \wedge B)$, due to the increased cognitive load associated with maintaining dual truth states (Anderson, 2013).

The challenge in tackling negation was also observed in the performance of the involved LLM models in our study. In logic proof construction, although DeepSeek-V3 (Table 3) achieved high accuracy on basic rules (e.g., Commutative (Com): 99.23 %), the model struggled with more complex rules (e.g., CP: 64.07 %). Our results suggest that LLMs encounter difficulties in propagating negation through inference chains, often struggling with expressions like $\neg(A \wedge B)$ and $\neg A \wedge \neg B$. This resembles a similar reasoning bottleneck found in human cognition.

The length of parent logic statements being combined also introduces complexity, usually via nested statements, which further introduces more cognitive load and working memory demands for humans. Similarly, a strong negative correlation was found between parent statement length and accuracy of LLMs. It may be that longer parent statements introduce ambiguities or combinatorial explosions that exceed the contextual reasoning capacity of current LLM models. Preprocessing inputs into simpler structures may improve the logical reasoning performance of current LLMs.

Overall, DeepSeek-V3 with FS_CoT achieved an average accuracy of 75 % for hint generation, but its hint accuracy was higher for earlier training problems and declined across the training levels as difficulty increased (from 87.9 % to 63.2 %). These accuracy levels are not high enough for integration into ITSs, as student trust can shift quickly when encountering even one that is inaccurate or not aligned with student thinking, potentially leading to help avoidance (Price, Zhi & Barnes, 2017). In our experiment, LLMs also sometimes generated sub-optimal hints based on the redundant statements derived by the students and failed to reorient the students to an optimal solution path.

Our LLM-generated hint explanations achieved higher scores on the consistency and clarity dimensions (Table 6), indicating that our selected LLM could follow specified rules and provide clear explanations. However, its justification and subgoaling scores were comparatively lower, suggesting LLMs sometimes struggle to articulate *why* a particular step was suggested within the larger problem or subgoal context. This limitation echoes critiques of LLMs as "stochastic parrots" (Bender et al., 2021) in educational contexts—they replicate surface-level patterns but lack intentional pedagogy, which is critical for learning (Aleven et al., 2003). Together with inaccuracy and these explanation ratings, our results highlight the need to build hybrid systems in which LLMs propose candidate hints, but they are checked for domain validity (e.g. symbolic checkers in math (He-Yueya et al., 2023)) and for pedagogical appropriateness.

## 6. Conclusion

Our work provides insights into the potential and current limitations of LLMs in logic tutoring systems, specifically examining their capabilities across proof construction, hint generation, and explanation quality. While models like DeepSeek-V3 and GPT-4o demonstrate promising performance in generating step-by-step solutions (achieving up to 86.7 % accuracy) and producing factually correct hints (75 % accuracy) with clear and consistent conceptual explanations, significant challenges remain in ensuring better pedagogical alignment. Our study also had several limitations. Our evaluation was limited to single-step accuracy for problem-solving in one domain. Future work should evaluate the optimality of LLM-constructed full solutions. Additionally, our study did not include reasoning models such as OpenAI's o1 or DeepSeek's reasoning models, which employ extended chain-of-thought processes and may offer improved performance in logical reasoning tasks. In the future, it can be investigated whether these reasoning models can provide more pedagogically sound explanations and proofs, as well as the feasibility of deploying such models at scale in classroom settings, considering their computational costs and response latencies. The human evaluation of hint explanations covered only 20 % of the dataset, potentially limiting generalizability; and although we evaluated 4 criteria for hint explanations, other pedagogical elements may be important for human understanding and learning. Future work should focus on developing hybrid architectures to combine LLM capabilities with learning science principles to bridge the gap between the potential of LLMs and the nuanced demands of a pedagogical environment.

## CRediT authorship contribution statement

**Sutapa Dey Tithi:** Writing – review & editing, Writing – original draft, Visualization, Validation, Project administration, Methodology, Investigation, Formal analysis, Data curation, Conceptualization. **Arun Kumar Ramesh:** Writing – review & editing, Software, Resources, Methodology, Conceptualization. **Clara DiMarco:** Writing – review & editing, Writing – original draft, Validation, Methodology, Conceptualization. **Xiaoyi Tian:** Writing – review & editing, Visualization, Validation, Supervision, Methodology, Investigation, Formal analysis, Conceptualization. **Nazia Alam:** Writing – review & editing, Validation, Resources, Data curation. **Kimia Fazeli:** Writing – review & editing, Validation, Resources, Data curation. **Tiffany Barnes:** Writing – review & editing, Validation, Supervision, Resources, Project administration, Methodology, Investigation, Funding acquisition, Conceptualization.

## Declaration of generative AI and AI-assisted technologies in the writing process

During the preparation of this work, the authors used OpenAI ChatGPT (GPT-4o) to improve language and readability. After using this tool/service, the authors reviewed and edited the content as needed and take full responsibility for the content of the publication.

## Declaration of competing interest

The authors declare that they have no known competing financial interests or personal relationships that could have appeared to influence the work reported in this paper.

## Acknowledgement

The work is supported by the National Science Foundation (NSF) grant 2013502.

## Appendix A. Few-shot prompting examples

This section provides two example problems used in few-shot prompting in our logic proof generation tasks.

*Example 1*

**Givens:** $(R * N) > O$, $-O * (-W > A)$, $(U > K) + (R * N)$, $K > -W$
**Conclusion:** $U > A$
**Step-by-step Proof:**

1. $(R * N) > O$, Given
2. $-O * (-W > A)$, Given
3. $(U > K) + (R * N)$, Given
4. $K > -W$, Given
5. $-O$, Simp (2)
6. $-W > A$, Simp (2)
7. $-(R * N)$, MT (5, 1)
8. $U > K$, DS (7, 3)
9. $U > -W$, HS (8, 4)
10. $U > A$, HS (9, 6)

*Example 2*

**Givens:** $A > E$, $H > (K * A)$, $-(K * E)$
**Conclusion:** $-H$
**Step-by-step Proof:**

1. $A > E$, Given
2. $H > (K * A)$, Given
3. $-(K * E)$, Given
4. $-K + -E$, DeM (3)
5. $-E + -K$, Com (4)
6. $E > -K$, Impl (5)
7. $A > -K$, HS (1, 6)
8. $-A + -K$, Impl (7)
9. $-(A * K)$, DeM (8)
10. $-(K * A)$, Com (9)
11. $-H$, MT (10, 2)